\documentclass{article}
\usepackage[a4paper]{geometry}

\usepackage{graphicx}%
\usepackage{multirow}%
\usepackage{amsmath,amssymb,amsfonts}%
\usepackage{mathrsfs}%
\usepackage[title]{appendix}%
\usepackage{xcolor}%
\usepackage{algorithm}%
\usepackage{algorithmicx}%
\usepackage{algpseudocode}%
\usepackage{listings}%
\usepackage{float}
\usepackage{caption}

\date{}

\providecommand{\keywords}[1]
{
  \small	
  \textbf{\textit{Keywords---}} #1
}
\begin{document}

\title{A Zero-Shot Reinforcement Learning Strategy for Autonomous Guidewire Navigation}

\author{Valentina Scarponi\footnote{MIMESIS-team, Inria de l'universit\'e de Loraine, MLMS team, Icube, univesit\'e de Strasbourg, France, \texttt{valentina.scarponi@inria.fr}}, Michel Duprez\footnote{MIMESIS-team, Inria de l'universit\'e de Loraine, MLMS team, Icube, univesit\'e de Strasbourg, France, \texttt{michel.duprez@inria.fr}}, Florent Nageotte\footnote{AVR team, Icube, univesit\'e de Strasbourg, France, \texttt{nageotte@unistra.fr}}~ and St\'ephane Cotin\footnote{MIMESIS-team, Inria de l'universit\'e de Loraine, MLMS team, Icube, univesit\'e de Strasbourg, France, \texttt{stephane.cotin@inria.fr} (corresponding author)}}







\maketitle

\abstract{
\noindent \textbf{Purpose}: The treatment of cardiovascular diseases requires complex and challenging navigation of a guidewire and catheter. This often leads to lengthy interventions during which the patient and clinician are exposed to X-ray radiation.
Deep Reinforcement Learning approaches have shown promise in learning this task and may be the key to automating catheter navigation during robotized interventions. Yet, existing training methods show limited capabilities at generalizing to unseen vascular anatomies, requiring to be retrained each time the geometry changes. 

\noindent \textbf{Methods}: In this paper, we propose a zero-shot learning strategy for three-dimensional autonomous endovascular navigation. Using a very small training set of branching patterns, our reinforcement learning algorithm is able to learn a control that can then be applied to unseen vascular anatomies without retraining.  

\noindent \textbf{Results}: We demonstrate our method on 4 different vascular systems, with an average success rate of 95\% at reaching random targets on these anatomies. Our strategy is also computationally efficient, allowing the training of our controller to be performed in only 2 hours.

\noindent \textbf{Conclusion}: Our training method proved its ability to navigate unseen geometries with different characteristics, thanks to a nearly shape-invariant observation space.
}

\keywords{Reinforcement Learning, Control, Endovascular Navigation, Robotics}



\section{Introduction}\label{sec1}
In the last decade, endovascular interventions have gained more and more importance for the treatment of a broad range of cardiovascular diseases, thanks to their minimal invasiveness, reduced costs and quick patient recovery \cite{vandenBerg, Sheela}. The treatment involves the navigation of guidewires and catheters inside the patient’s vessels until reaching the target, which can be located in different districts of the body, such as the brain, the heart, or the liver. Then various devices, such as balloons, stents, or coils can be deployed to locally treat the pathology. Each of these steps is performed under fluoroscopic imaging, which provides only a limited two-dimensional projective view of the anatomy.  

Efficiently and safely navigating the catheter and guidewire through the vascular anatomy is essential to ensure minimal exposure of the patient and clinician to X-ray radiations induced by the fluoroscopic imaging system. This requires a perfect knowledge of the anatomy, excellent control of the device, and a deep understanding of the fluoroscopic visualization. Yet, even experienced clinicians can spend tens of minutes until reaching certain targets. A possible improvement can be offered by robotic systems \cite{Puschel2022}. However, these robots remain master-follower systems that manipulate the devices following the inputs provided by the clinician. To provide the clinician with more support, current research works are now focusing on the development of autonomous and semi-autonomous systems. To this end, several deep reinforcement learning (DRL)-based methods have been proposed. They generally take fluoroscopic images as input and predict a control action (rotation and translation) to be applied at the proximal end of the device. Some works perform the training and apply the learned control in entirely simulated environments \cite{Tian2023}. Others train the neural controller with images of the phantom in which the navigation is later performed \cite{Wang22, Kweon} or perform the training in simulated environments and then feed the neural controller with images of the phantom during the navigation \cite{Push}. 

The current limitation of these different works is twofold. First, they rely on fluoroscopic images as input of the neural network, which is known to lead to ambiguities about the tip orientation, and therefore incorrect controls can be predicted. Second, the training is not generic, requiring a new training for each new patient. The learned control is also generally not compatible with a deformation of the anatomy. The development of neural controllers capable of realizing tasks not only in environments they have already explored but also in situations they have never seen before \cite{Miranda_2023, Kirk_2023} remains an important challenge in DRL. To address this issue in the context of endovascular interventions, Kweon \textit{et al.} \cite{Kweon} proposed a segment-wise learning approach to accelerate training using human demonstrations, transfer learning, and weight initialization. However, this approach still requires to train the network each time the environment is expanded or changed.
In the work proposed by Karstensen \textit{et al.} \cite{Karstensen}, the authors exploited recurrent neural networks to design a learning-based controller able to navigate various geometries. After 3.5 million exploration steps on various aortic arches, the controller reports a 75\% success rate in navigating different aortic arches, which reduces to 29\% when vessels from a real patient are used. Furthermore, the method requires a different training for each vessel structure, which limits the applicability of the system when various structures need to be navigated. Chi \textit{et al.} \cite{Chi2020} implemented a generative adversarial imitation learning method aimed at learning to perform the catheterization of the brachiocephalic artery (BCA) and left common carotid artery (LCCA). In this work, they trained the network using expert demonstration on a reproduction of the aortic arch. On the training geometry, the system reported a success rate of 94.4\% for the BCA cannulation and of 88.9\% for the LCCA cannulation, but the performances were reduced to 69\% for the BCA and 72.2\% for the LCCA when the aortic type was changed. The same author proposed in a previous work \cite{Chi2018} a generalization method  that did not involve the use of deep neural networks: using human demonstration, they trained a statistical model to perform the cannulation of the innominate aorta on a 3D model of the type I aortic arch and applied the same controller to variations of the same aortic arch type. This technique reported an average 98\% success rate in new but very similar geometries, using human demonstrations for each new task.

In this paper, we present a DRL method able to generalize to any vascular anatomy, assuming it is possible to navigate toward different target areas using the same guidewire tip shape. As a side effect, our method is also able to handle shape variations due to physiological movements (e.g., breathing, cardiac motion). To avoid ambiguities due to 2D shape reconstruction from single plane fluoroscopic images, we propose to rely on shape sensing technologies based on FBGs to obtain a 3D estimation of the tip shape and orientation of the guidewire. Another possible method to obtain this 3D shape during endovascular interventions is the one proposed by Chen \textit{et al.} \cite{Chen20}, who exploited ultrasound imaging. Using our DRL method, we demonstrate a success rate of almost 100\% in several 3D complex anatomies, and a very robust behavior to deformations of the vascular tree. 

Section~\ref{sec:method} presents the main contributions of our method, in particular the representation of the observation space that allows for an invariance to rigid transformations and small deformations. The rest of the article is organized as follows: section~\ref{sec:Results} presents our results on different synthetic test cases, and highlights the generalization capability of our DRL method and section~\ref{sec:Conclusion} presents our plan for further improving this approach and testing it on vascular phantoms. \\Please refer to \textit{Online Resource1} for a video summarizing both the method and the results presented in this work.


\section{Materials and Methods}
\label{sec:method}

\begin{figure}[ht!]
	\centering
 \captionsetup{width=.8\linewidth}
	\includegraphics[width=0.9\columnwidth]{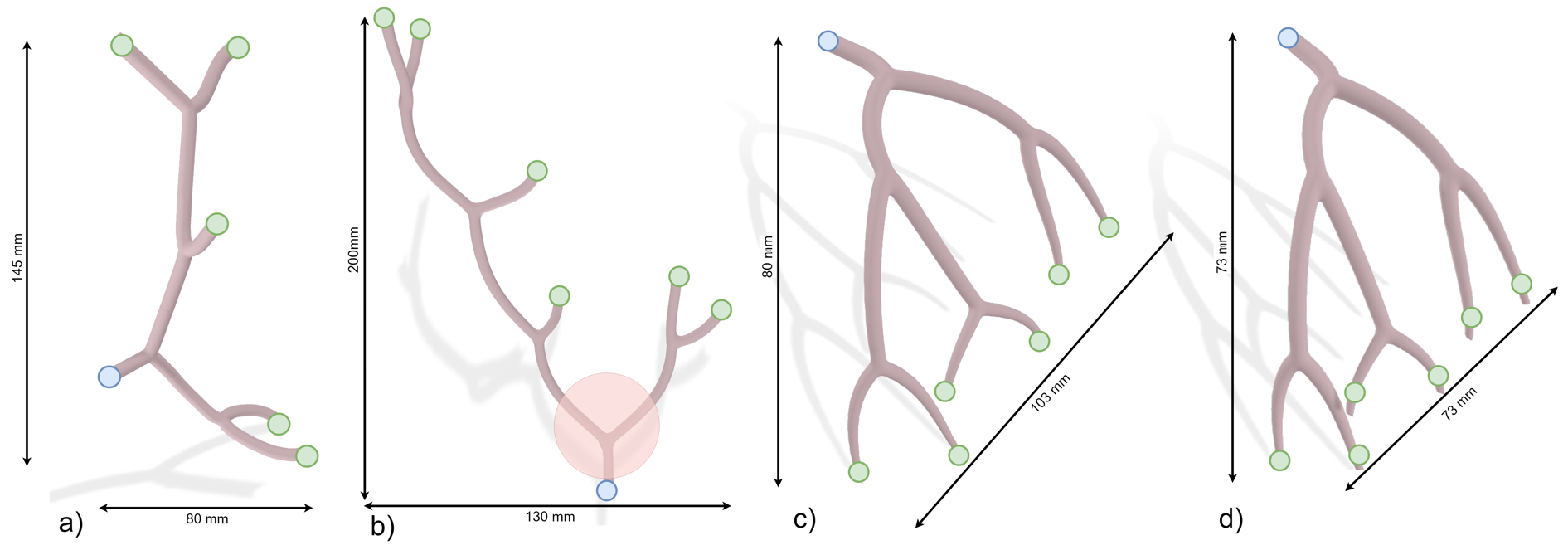}
	\caption{Geometries used to test the algorithm. The blue dots represent the insertion points, while the green dots show all possible target locations. The red circle shows an example of a bifurcation region.
    }
    \label{fig:geometries}
\end{figure}

Our objective is to develop a \textit{generalized neural controller} able to continuously control the rotation of a guidewire through a complex vascular tree, from its insertion point until one of the predetermined targets is reached (see Figure~\ref{fig:geometries}). This control is performed at the proximal end of the device, and accounts for its deformation during navigation.

The core of our method is a \textit{nearly shape-invariant observation space}, allowing to train once on a set of bifurcation patterns, and then performing the control on a complete and unseen vascular tree (see section~\ref{sec:testAnatomies}).
For this method to be effective, it is crucial to choose an algorithm that encourages the exploration of the environment, to ensure that the agent visits in each anatomy all possible states. To this purpose, we adopted the Soft Actor-Critic (SAC) algorithm  \cite{Haarnoja}, a state-of-the-art off-policy DRL algorithm that introduces an entropy term to enhance exploration (see section~\ref{sec:SAC}).

Another important objective of our work is the possibility of navigating 3D non-planar anatomies.
Three-dimensional anatomical models can be easily retrieved from MRI or CT preoperative scans. To reconstruct the 3D shape of the device, recent developments of Fiber Bragg Gratings (FBG) sensors \cite{Othonos}, which are modified optical fibers, make it possible to reconstruct the 3D shape of guidewires or catheters in which they can be embedded. Our simulated environment builds upon this choice, and uses accurate Finite element simulations to facilitate sim-to-real transfer (see section~\ref{sec:sim_env}).


\subsection{Training environment}
\label{sec:training_env}

\subsubsection{Simulation of guidewire nagivation}
\label{sec:sim_env}
The training of DRL algorithms requires a large number of interactions between the agent and the environment. In this work, to evaluate the effectiveness of our method while avoiding the costs associated with the use of a real experimental setup, 
the training was carried out in a simulated environment, modeling both the physics of the device and its interactions with the vessel walls. An accurate modelization of the physics is crucial for the future transposition of the controller on a real setup. Currently available simulators, such as CathSim \cite{Cathsim}, control the device from its tip, which does not reproduce the control of passive instruments which can only be manipulated from their proximal end. We therefore built our own Simulator, based on the open source SOFA framework \cite{Sofa}. In this framework, we rely on a model based on Timoshenko beam theory \cite{Bitar2015} to simulate the deformation of the guidewire \cite{Meng}. The system to be solved is reported in Equation \eqref{eq:1} in its matrix form.
\begin{equation}
(\textbf{M} - dt^2 \textbf{K})\Delta v = dt\cdot f(x(t))+dt^2 \cdot \textbf{K}v(t)
\label{eq:1}
\end{equation}
$\textbf{M}$ represents the mass matrix, $\textbf{K}$ the stiffness matrix and $dt$ the time step. $v$ and $\Delta v$ denote the velocity and the velocity variation respectively and $f$, which is a function of the current positions $x(t)$, represents the internal and external forces applied to the system. The tridiagonal nature of the matrix $(\textbf{M} - dt^2 \textbf{K})$, which is a consequence of the structure of the guidewire model, allows the use of a Block Tridiagonal (BTD) linear solver for its inversion, which helps reduce the computational time to only a few milliseconds per time step. 
\\The interactions between the vessel wall and the guidewire are computed by first solving the dynamics of the guidewire (vessels are supposed rigid and static) and then performing a collision detection. Then, instead of using a penalty-based method for solving the contact problem, a Lagrange Multiplier approach and a single linearization by time step are adopted. This ensures more stability on the simulation. Equation \eqref{eq:1} then becomes:
\begin{equation}
(\textbf{M} + dt\frac{df}{d\dot{x}}+dt^2\frac{df}{dx}) \Delta v =
-dt(f+dt\frac{df}{dx}v)+dt \textbf{H}^T\lambda,
\label{eq:2}
\end{equation}
where $\textbf{H}^T \lambda$  is the vector of constraint forces contribution, with $\textbf{H}$ containing the constraint directions arising from the collision detection, and $\lambda$ the Lagrange multipliers.
\\The physics of the guidewire model is then corrected by computing the contact force $\lambda$ using a Gauss-Seidel algorithm.

As a result, it is possible to simulate the navigation of the virtual device at an average of 90 frames per second, thus allowing for relatively fast training times (see section~\ref{sec:Results}). In addition, we compared our simulation to an actual guidewire in a rigid phantom that was scanned and reconstructed in 3D. We noticed an average error of $2.0 \pm 0.9 \; mm$ which shows that our simulation has the potential to transfer the learned policy to the real-world domain.  


\subsection{Training strategy}
\label{sec:training}

\subsubsection{Soft Actor-Critic algorithm}
\label{sec:SAC}
Reinforcement learning (RL) is a sub-field of machine learning in which an agent interacts with an environment and learns how to reach a specific goal. The problem is usually formulated as a Markov Decision Process \cite{Bellman}, \textit{i.e.} as a tuple $(\mathcal{S,A},P,r,\gamma)$, where $\mathcal{S}$ and $\mathcal{A}$ are a set of states and actions respectively. The environment is (often only partially) described through the \textit{observation space} $\Omega$. At each time step, an action $a_t \in \mathcal{A}$ is chosen by a policy $\pi$ that maps states to actions: $\mathcal{S}\rightarrow\mathcal{A}$. This action makes the system transition from the state $s_t \in \mathcal{S}$ to the state $s_{t+1} \in \mathcal{S}$.  $P(s_{t+1}|s_t,a_t)$ is the probability density of the next state $s_{t+1} \in \mathcal{S}$ given the current state $s_{t} \in \mathcal{S}$ and action $a_{t} \in \mathcal{A}$ and $r(a_t,s_t)$ is the reward obtained by the agent for this transition. 

In recent years, many variations of the classical formulation of the DRL problem have been developed. Particular interest has been raised by the SAC algorithm, which is an off-policy DRL algorithm that outperforms several previous methods \cite{Haarnoja}, in particular deep deterministic policy gradient (DDPG), which are very often used in autonomous catheter navigation \cite{Tian2023, Push}.
The SAC algorithm optimizes the following objective function: 
\begin{equation}
    J(\pi)=\sum_{t=0}^T\mathbb{E}_{(s_t,a_t)\sim \rho_\pi}[r(s_t,a_t)+\alpha \mathcal{H}(\pi(\cdot|s_t))]
    \label{eq:obj_fun}
\end{equation}
in which $\rho_\pi(s_t)$ and $\rho_\pi(s_t,a_t)$ denote the state and state-action marginals of the trajectory distribution induced by a policy $\pi(a_t|s_t)$. With respect to the usual formulation of RL's objective functions, SAC introduces the entropy term $\mathcal{H}(\pi(\cdot|s_t))$ 
that encourages the exploration of the environment, avoiding repeatedly choosing an action that may exploit inconsistencies in the approximated Q-function. By choosing an appropriate value of $\alpha$ we can help the agent generalize its learned policy, which is important given our objective.

\subsubsection{Zero-shot Reinforcement Learning Strategy} 
\label{sec:learning_strategy}
While the choice of the DRL method (and hyper-parameters) plays an important role in our solution, the definition of the observation space, and selection of the training geometries, will prove essential to learn a generalizable control. Unlike previous works, we do not train our network on the test anatomy. Instead, we train it on a set of bifurcation patterns, not directly related to the test anatomies. We process the 3D shapes and compute their centerlines for both the test anatomy and training shapes. This computation is done automatically from the 3D vascular geometry. Then, at the beginning of each training episode, a target is chosen among a set of possible final states, typically (but not necessarily) placed near the end of each terminating branch. Then, a unique path connecting the starting point to the target location is computed from the centerlines of the training anatomy. For each episode, the starting rotation of the guidewire around its axis and its orientation with respect to the centerline are chosen randomly to enhance the exploration of the environment. The controller is set to push the device at a constant speed during the whole episode.
The parameters of the DRL algorithm and details on the choice of the training geometries will be given in section ~\ref{sec:observation_space} and ~\ref{sec:training_geoms} respectively.

 \begin{figure}[h!]
	\centering
 \captionsetup{width=.8\linewidth}
	\includegraphics[width=0.8\columnwidth]{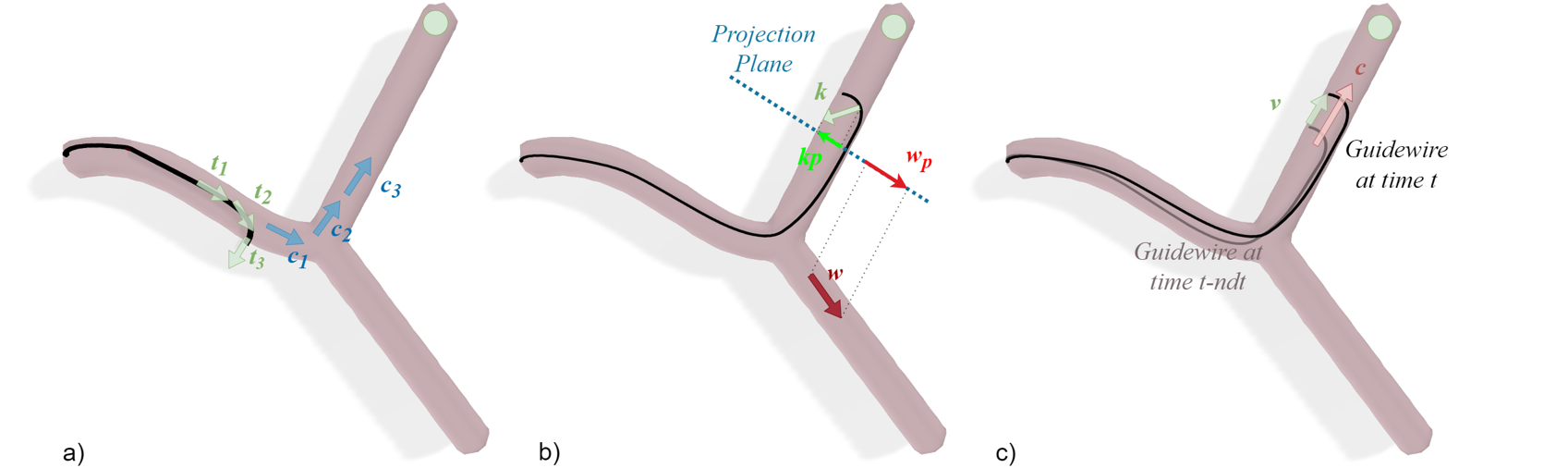}
	\caption{Our observation space is composed of: 1) $t_i \cdot c_j$, with $i \in [1;3] \in \mathbb{Z}$ and $j \in [1;3]\in \mathbb{Z}$ (a), 2) the normalized distance between the tip of the guidewire and the target, 3) the chosen action, 4) $k_p \cdot w_{p}$ (b) 5) $v \cdot c_i$ (c).}
    \label{fig:obs_space}
\end{figure}
\subsubsection{Nearly shape-invariant observation space}
\label{sec:observation_space}

To enforce generalization of the learned control, we designed an observation space that is rotation and translation invariant, but also shows little sensitivity to the shape variation of the bifurcation. The observation space allows the agent to build its internal representation of the state, which the controller uses to make its decision about the best action. By letting the agent build this representation from observations that describe its \textit{relative} position to the environment, we can make it robust to a variety of transformations of the bifurcation shape. To the end, we propose to construct the observation space $\Omega$ as follows:
$$ \Omega = \{\boldsymbol{\theta}_t, \boldsymbol{\theta}_{t-ndt}, \lambda_{t}, \lambda_{{t-ndt}}, a_t, \omega, s\}$$ 

\begin{itemize}
\item Let $\mathbf{t}_i, i \in [1,N]$ be the tangent vector at the coordinate $\mathbf{x}_i$ along the tip of the guidewire, and $\mathbf{c}_j, j \in [1,N]$ the tangent vector of the centerline at position $\mathbf{x}_j = \mathbf{x}_i+\mathbf{h}$. We define $\theta_{i} = \mathbf{t_i}\cdot \mathbf{c_i}$ $ \forall \; i \in [1,N] \in \mathbb{Z}$. 
\item We then define $\boldsymbol{\theta}_m=[\theta_1,\theta_2,...,\theta_N]_m$, with $m \in \{t; t-ndt\}$.
\item $\lambda_{t}$ and $\lambda_{t-ndt}$ represent the normalized distance between the tip of the guidewire and the target at time $t$ and $t-ndt$
\item $a_t$ is the action that determines the transition of the system from $s_{t-ndt}$ to $s_t$
\item $\omega = \mathbf{k}_p \cdot \mathbf{w}_p$, where  $\mathbf{k}_p$ is the projection of the curvature vector $\mathbf{k}$ of the tip and $\mathbf{w}_p$ is the projection of the vector describing the direction of the wrong branch $\mathbf{w}$ onto a plane perpendicular to the centerline of the branch leading to the target (Fig. \ref{fig:obs_space}b).
\item $s=\mathbf{v} \cdot \mathbf{c}$, where $\mathbf{v}$ describes the current velocity of the guidewire and $\mathbf{c}$ the tangent to the centerline near the tip of the guidewire (Fig. \ref{fig:obs_space}c).
\end{itemize}

%
It is important to notice that all the parameters used to build the observation space can be computed in both the virtual (training) environment and in a real setup. The metrics involving the vessel geometry can be retrieved from pre-operative MR or CT images, while the parameters describing the tip shape of the guidewire can be determined after shape reconstruction of the FBG data \cite{Al-Ahmad}. To reduce the effect of potential reconstruction errors, we rely on multiple grating measurements to define the tip orientation ($\mathbf{t}_i$).

Another key element to learning the optimal action is the engineering of the reward function. We design our reward function as the weighted sum of three terms (Eq. \eqref{eq:rewardFunction}):
\begin{equation}
    r(s_t, a_t) = \underbrace{\frac{2}{1+e^{5(\omega-0.1)}}-1}_{a} + \underbrace{0.5\left(1-\frac{\lambda_{t}}{\lambda_{0}}\right)}_{b} + \underbrace{(-0.2|a_t|)}_{c}
    \label{eq:rewardFunction}
\end{equation}
where $\lambda_t$ and $\lambda_0$ are respectively the current and the initial distance between the tip and the target. $a$ encourages the agent to obtain a tip direction ($\mathbf{k}_p$, Figure \ref{fig:obs_space}) opposite to $\mathbf{w}_p$. This function is a modified version of the sigmoid activation function, largely used in the deep learning field. In this modified version, the output $\gamma$ is a decreasing function with $\gamma \in [-1;1] \in \mathbb{R}$. $b$ increases as the target is approached, while the term $c$ discourages the agent from rotating the instrument when it is unnecessary.

\subsubsection{Design of the training anatomies}
\label{sec:training_geoms}
 \begin{figure}[H]
	\centering
 \captionsetup{width=.8\linewidth}
	\includegraphics[width=.9\columnwidth]{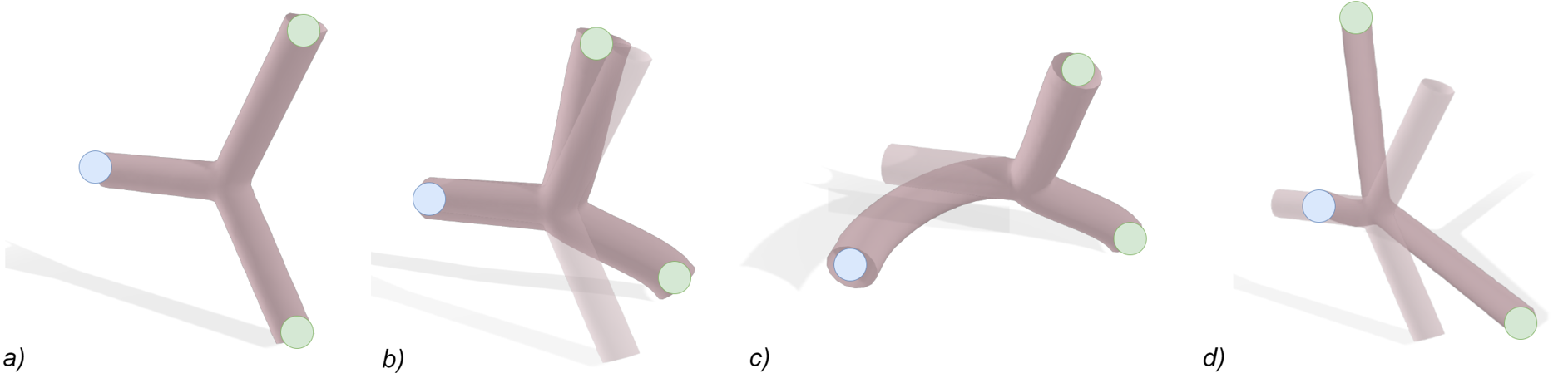}
	\caption{Geometries used to study the sensitivity of our training with respect to changes in position (a), orientation (a) and shape (b,c,d). The blue dot marks the entry branch, the green ones the exit.}
    \label{fig:geom_choice}
\end{figure}
To choose a proper dataset for the training, we studied the sensitivity of our method to the orientation, the position and the shape of the geometry. We trained our network on a single bifurcation (Fig. \ref{fig:geom_choice}a) and tested it in 5 variants of this bifurcation: a rotation of the model, a translation, a variation (var$_1$) of the shape of the exit branches (Fig. \ref{fig:geom_choice}b), a variation (var$_2$) of both the exit and entry branch shapes (Fig. \ref{fig:geom_choice}c) and an important variation (var$_3$) of the exit branches shape (Fig. \ref{fig:geom_choice}d).

Once trained, we first assessed that the agent was able to get a 100\% success rate in the training anatomy. Then, the agent was required to perform the navigation 50 times in each of the 5 scenarios: 3 shapes variations, rotation and translation of the training shape. The success rates were $100\%$ (rotation), $100\%$ (translation), $100\%$ (var$_1$), $62\%$ (var$_2$) and $86\%$ (var$_3$), showing that our training is mostly sensitive to variations of the entry branch shape.

Therefore, and as we will illustrate in section~\ref{sec:Results}, we can already anticipate that with only a few training shapes, we can maximize the chances of generalizing the learning to various complete anatomies. 

\section{Results} 
\label{sec:Results}
 \begin{figure}[t]
	\centering
 \captionsetup{width=.8\linewidth}
\includegraphics[width=0.9\columnwidth]{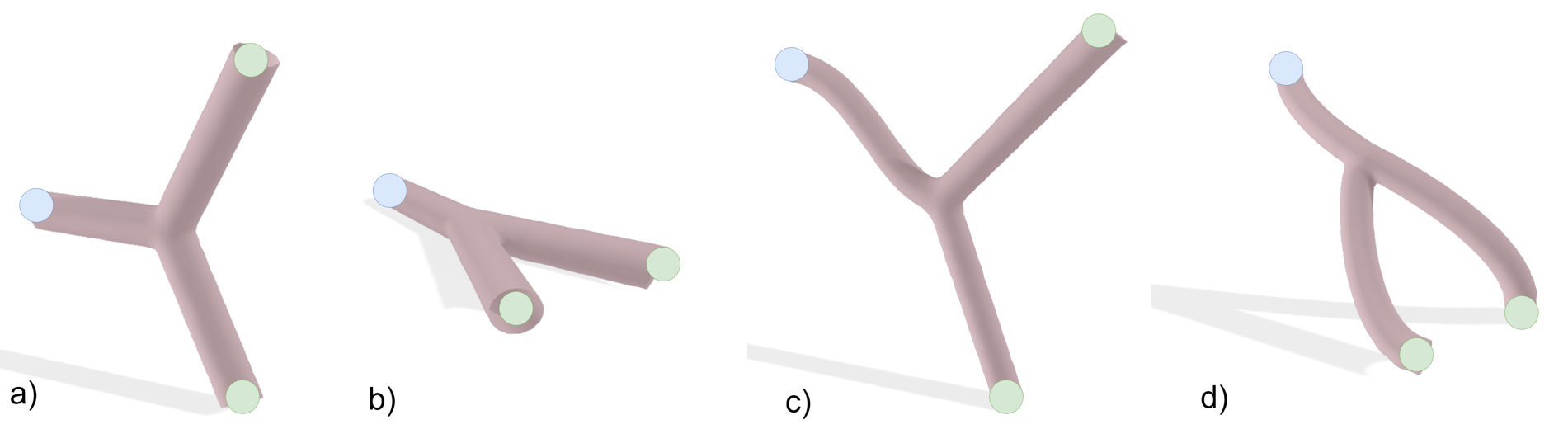}
	\caption{Geometries chosen to train the agent and to cover the whole observation space. In each model, the blue point represents the starting point, and the green one the target location.}
    \label{fig:geom_train}
\end{figure}
This section presents the training dataset (section \ref{sec:traininganatomiesRes}) constituting the environment in which the SAC agent learned to control the guidewire following our training strategy. The controller was then used to navigate various complex unseen anatomies (section \ref{sec:testAnatomies}) to test its generalization capabilities. The guidewire selected for this work presents a $6 \; mm$ curved tip characterized by a curvature of $0.2 \; \frac{1}{mm}$ and a total length of $310 \; mm$. Both the training and the tests were performed on a computer equipped with 32 GB of RAM, a 24-core Intel(R) Core(TM) i7 running at $3.40$ GHz and an NVIDIA GeForce GTX 1080 GPU. In this work, the SAC implementation available in Stable Baselines3 library \cite{stable-baselines3} was used, while the virtual environment was developed using SOFA \cite{Sofa}.

\subsection{Training anatomies}
\label{sec:traininganatomiesRes}
For the purpose of this work, a training dataset composed of only 4 constant radius anatomies was designed (Fig. \ref{fig:geom_train}). This was identified as the minimum dataset to respect the criteria described in section \ref{sec:training_geoms}: each geometry presents different characteristics in terms of either the shape of the access vessel or the shape of the exit branch, allowing the agent to explore the shape variation observed in real vascular trees. Using this dataset, the agent was trained until the convergence, obtained after 120,000 time steps, equivalent to about 130 episodes. Both the actor and the critic networks are composed of three 256-neuron layers. After tuning the hyperparameters, a learning rate of $10^{-4}$, a batch size of $256$, a buffer size of 10,000 and a discount factor of 0.98 were set, while the entropy coefficient was learned autonomously during the training. 1,000 steps were performed before starting the training of the model, which was updated at every time step.

At the beginning of each episode, one of the 4 anatomies was randomly chosen, along with the target location (green points in Fig. \ref{fig:geom_train}) and the guidewire was placed at the entrance of the selected vessel (blue points in Fig. \ref{fig:geom_train}) with a random orientation. The time required to complete the training was about 2 hours.
The control learned by the controller was first tested on the training anatomies, under the same conditions used for the training itself. The navigation was performed on a total of 100 episodes and the controller obtained a success rate of 95\%.

\subsection{Navigation test on complex vascular trees}
\label{sec:testAnatomies}
To test the efficacy of our method, the trained controller was used to navigate complex anatomies composed of various connected bifurcations. To this purpose, three different geometries were considered. The first anatomy was obtained by composing the training geometries (Fig. \ref{fig:geometries}a) and aimed at demonstrating that learning on a single set of bifurcations can lead to a correct control on a combination of these bifurcations. The second anatomy has a constant vessel diameter, and contains up to 4 subsequent unseen bifurcations (Fig. \ref{fig:geometries}b), while the third one is a reconstruction of the left coronary arteries, characterized by unseen bifurcation shapes and variable vessel radii (Fig. \ref{fig:geometries}c). For each anatomy, the controller was tested for 50 episodes, each starting with the guidewire placed at the insertion point of the anatomy (blue points in Fig. \ref{fig:geometries}) with a random initial orientation. An episode was considered finished either when the target was reached (green points in Fig. \ref{fig:geometries}, \textit{successful episode}) or when, navigating a bifurcation, the wrong branch was chosen (\textit{failure}). In these complex anatomies, the control of the guidewire was entrusted to the agent only in the proximity of the bifurcation (\textit{i.e.}, a region comprising an area of radius $20\;mm$ and centered at the center of the bifurcation): indeed, outside these regions, the control is trivial and only consists of pushing forward the device. Every time a bifurcation was reached, an intermediate target was inserted in the branch that needed to be navigated to reach the final target. In the anatomy composed of the training geometries, the controller obtained a success rate of 98\%. The success rate in the geometry composed of unseen bifurcations was 100\%, and 94\% in the coronaries-like anatomy. 

We then performed an additional test, on a deformed version of the coronary arteries (Fig. \ref{fig:geometries}d). This deformation corresponds to the maximum deformation amplitude observed during a cardiac motion. On that new unseen anatomy, our neural controller succeeded 45 times out of 50 trials to reach random targets (i.e. a 90\% success rate).


\section{Conclusion}\label{sec:Conclusion}
In this work, we developed an efficient method to train a neural controller able to perform navigation tasks on unseen complex anatomies, after being trained on a set of only 4 bifurcation geometries. We obtained an average 95\% success rate when navigating new anatomies, outperforming previous works that reported a success rate of 72\% \cite{Chi2020} and 75\% \cite{Karstensen} when trying to generalize the task to unseen models. 

Exploiting further our zero-shot reinforcement learning strategy, our very next step will consist of navigating the coronary arteries while the (simulated) heart is beating. Our results on navigating the 2 deformed coronary shapes are very encouraging, but additional work is required to achieve this result.


Finally, the simulated environment used in this work allows
to accurately simulate the physics of real catheters and guidewires. Yet, for sim-to-real transfer of the learned control, additional fine-tuning of the physical and geometrical characteristics of the devices (diameter, Young Modulus, etc.) is needed and the
simulation of blood flow would be a plus. We have already started to work in this direction, motivated by previous works \cite{Push} that used a similar virtual environment and successfully transferred the learned control to navigating a vascular phantom.
\section*{Declarations}
\textbf{Funding:} This work of the Interdisciplinary Thematic Instiute HealthTech, as part of the ITI 2021-2028 program of the University of Strasbourg, CNRS and Inserm, was supported by IdEx Unistra (ANR-10-IDEX-0002) and SFRI (STRAT’US project, ANR-20-SFRI-0012) under the framework of the French Investments for the Future Program.
\\\textbf{Financial Interests:} The authors have no relevant financial or non-financial interests to disclose.

\bibliography{bibliography}
 \bibliographystyle{unsrt}

\end{document}